\documentclass[10pt]{article} 

\usepackage[preprint]{rlj} 

\usepackage{amssymb}            
\usepackage{mathtools}          
\usepackage{mathrsfs}           
\usepackage{graphicx}           
\usepackage{subcaption}         
\usepackage[space]{grffile}     
\usepackage{url}                
\usepackage{lipsum}             

\usepackage[disable]{todonotes} 
\newcommand{\HA}[1]{\todo[color=lightgray,inline]{HA: #1}}

\title{Environment Agnostic Goal-Conditioning, \\
      A Study of Reward-Free Autonomous Learning}

\setrunningtitle{Environment Agnostic Goal-Conditioning}

\author{Hampus Åström\textsuperscript{1},
Elin Anna Topp\textsuperscript{1},
Jacek Malec\textsuperscript{1}}
    
\emails{\{hampus.astrom,elin\_anna.topp,jacek.malec\}@cs.lth.se}

\affiliations{
$^{1}$\textbf{Dept. of Computer Science, Lund University, Lund, Sweden}
\par 
}

\contribution{
    We apply goal-conditioning on non-goal environments, and show that ignoring external rewards and training guided by goal selection can yield comparable results to an externally guided baseline.
    }
    {
    Prior work introduced goal-conditioning and agents that self-select goals \citep{ andrychowiczHindsightExperienceReplay2018, colasAutotelicAgentsIntrinsically2022}.
    }

\contribution{
    We present a study on the success rates of multiple goals, during training with autonomous goal conditioning, in three different environments.
    We show that, while the success rate for individual goals varies significantly during training, the average goal success rate gradually improves and stabilizes.
    }
    {
    In previous work, \citet{florensaAutomaticGoalGeneration2018} study goal coverage, as an overall measure of average goal success rate, without explicitly measuring the performance of individual goals over time.
    }

\contribution{
    We provide an extension to Stable-Baselines3 \citep{stable-baselines3}, which enables applying goal conditioning to non-goal environments via a wrapper, with modular goal evaluation and selection. \footnote{\url{https://github.com/HampusAstrom/goal-exploration}}
    }
    {
    Our code contribution relies on and merely extends the preexisting repository Stable-Baselines3 \citep{stable-baselines3}.
    }


\summary{In this paper we study how transforming regular reinforcement learning environments into goal-conditioned environments can let agents learn to solve tasks autonomously and reward-free.
We show that an agent can learn to solve tasks by selecting its own goals in an environment-agnostic way, at training times comparable to externally guided reinforcement learning.
Our method is independent of the underlying off-policy learning algorithm.
Since our method is environment-agnostic, the agent does not value any goals higher than others, leading to instability in performance for individual goals.
However, in our experiments, we show that the average goal success rate improves and stabilizes.
An agent trained with this method can be instructed to seek any observations made in the environment, enabling generic training of agents prior to specific use cases.
}

\begin{document}

\makeCover  
\maketitle  





\begin{abstract}
In this paper we study how transforming regular reinforcement learning environments into goal-conditioned environments can let agents learn to solve tasks autonomously and reward-free.
We show that an agent can learn to solve tasks by selecting its own goals in an environment-agnostic way, at training times comparable to externally guided reinforcement learning.
Our method is independent of the underlying off-policy learning algorithm.
Since our method is environment-agnostic, the agent does not value any goals higher than others, leading to instability in performance for individual goals.
However, in our experiments, we show that the average goal success rate improves and stabilizes.
An agent trained with this method can be instructed to seek any observations made in the environment, enabling generic training of agents prior to specific use cases.
\end{abstract}


\section*{Introduction}

Reinforcement learning (RL) is a good step towards the idea of machines that learn autonomously. RL only needs to provide the agent with an environment with which to interact, not an expensive trove of data. 
However, defining a reward function is problematic, as it often needs to be hand-made by an engineer to guide the agent to optimal behavior. 
Ideally, an autonomous agent would learn meaningful behavior without the external guidance of a problem-specific reward function.
We propose that by wrapping environments in a goal-conditioning framework, and training without any external rewards to reach points in the environment, we can create agents that learn broad capabilities that can be accessed for specific tasks. 



\subsection*{Contributions}


\begin{itemize}
  \item We apply goal-conditioning on non-goal environments, and show that ignoring external rewards and training guided by goal selection can yield results comparable to an externally guided baseline. This expands on prior work that introduced goal-conditioning \citep{andrychowiczHindsightExperienceReplay2018} and agents that self-select goals \citep{ colasAutotelicAgentsIntrinsically2022}.
  \item We present a study on the success rates of multiple goals, during training with autonomous goal conditioning, in three different environments. We show that, while the success rate for individual goals varies significantly during training, the average goal success rate gradually improves and stabilizes. In previous work, \citet{florensaAutomaticGoalGeneration2018} study goal coverage, as a measure of average goal success rate, but without explicitly measuring the performance of individual goals over time.
  \item We provide an extension to Stable-Baselines3 \citep{stable-baselines3}, which enables applying goal conditioning to non-goal environments via a wrapper, with modular goal evaluation and selection.\footnote{\url{https://github.com/HampusAstrom/goal-exploration}}
\end{itemize}

\section*{Related Works}



In regular reinforcement learning, a predefined reward function describes what behavior is desirable. This most often means that there is a single optimal policy. Even when well-defined external rewards exist, they can be sparse and deceptive. During learning, an agent often cannot rely solely on these signals to guide its actions. \citet{pathakCuriosityDrivenExplorationSelfSupervised2017} suggest solving this with \textit{Intrinsic Curiosity}, a method that treats errors in a transition model as intrinsic rewards to guide exploration towards interesting states. Although efficient, this relies on distorting the agent's perception and can suffer from problems with detachment and derailment \citep{ecoffetFirstReturnThen2021}. 
Exploration can also be done with the help of goals \citep{florensaAutomaticGoalGeneration2018, ecoffetFirstReturnThen2021, colasAutotelicAgentsIntrinsically2022}.


Goal-conditioned reinforcement learning \citep{andrychowiczHindsightExperienceReplay2018} proposes that we describe to the agent what we want it to achieve in each episode. This can be done by adding a goal description to the observations and providing a goal selection method and a goal evaluation scheme. 
This is particularly valuable if one wants to learn different or parameterized tasks, as well as in autonomous learning \citep{colasAutotelicAgentsIntrinsically2022,colasCURIOUSIntrinsicallyMotivated2019,florensaAutomaticGoalGeneration2018,liuGoalConditionedReinforcementLearning2022a}.
If goal-conditioning can be applied in an environment-agnostic way, agents can learn autonomously and reward-free, in unknown environments.

\section*{Method}


This paper explores whether and how goal conditioning with self-selection of goals can be formulated in a generic way. This approach is evaluated by training agents with a goal wrapper in an environment-agnostic, reward-free formulation and comparing them to regular RL methods that are aware of external goals. Agents trained in such a way can be supplied with tasks via the goal formulation, post-training, enabling flexible agents that can do various tasks in the environment they trained on. 

We have created a wrapper in the Stable-Baselines3 framework \citep{stable-baselines3}, to enable goal-conditioning for non-goal RL environments. It takes observations from the underlying environment, together with goals, and uses them to evaluate when goals have been achieved.

Goal success evaluation is modular; in our experiments we have used normalized distance to determine goal success in continuous environments and exact matching in discrete environments. We terminate when the agent reaches a goal.

Goal selection is also modular, and we present three different example methods that are used to select goals for the agent during training: uniform sampling, novelty selection, and intermediate difficulty selection. 
Goals should prioritize exploration, while making sure that the agent can improve its capacity to reach found goals and avoid forgetting previous capabilities. In this way, goal selection selectively collects training data. In future work, our aim is to explore supplementing goal selection with smarter filling and sampling from the replay buffer to get more efficient and stable goal learning.

We constrain our study to single observation goals, rather than goal trajectories, as goal conditioning already expands the task considerably. With external rewards, there is often a single optimal strategy, but with goal-conditioning, each goal might need a unique strategy. Hindsight Experience Replay \citep{andrychowiczHindsightExperienceReplay2018} reduces issues with the expanding scope, by enabling learning from every episode, by the principle that each episode shows a way to reach the observations in that episode. We use off-policy learning, and since we only present studies on environments with discrete action spaces here, we use \citet{stable-baselines3}'s implementation of Deep Q-Networks (DQN) \citep{mnih2013playingatarideepreinforcement} (though we have confirmed that our method works with Soft Actor-Critic \citep{haarnoja2018softactorcriticoffpolicymaximum} as well).








\subsection*{Goal selection strategies}

In the general case, a goal selection strategy might want to take into account the current state or observation, especially when starting conditions change drastically or when goals are re-selected during an episode. There are only minor variations in start conditions of the environments of our trials here, and we terminate whenever a goal is achieved, so in this work that factor is ignored when selecting goals. 

We use three main methods to select goals: Uniformly random, novelty seeking, and intermediate success rate selection (the latter inspired by intermediate difficulty goal selection \citep{florensaAutomaticGoalGeneration2018}. With all methods, we add some uniform random selection to avoid getting stuck in a local goal subset.

\subsubsection*{Uniform random}

The simplest way to select goals is to randomly sample over the observation space uniformly. In environments where only a fraction of the observation space are viable, reachable observations, this method risks selecting a large fraction of nonviable goals. 

\subsubsection*{Novelty selection}

Our novelty selection method selects goals by valuing less visited areas higher. In environments with discrete observations, this is simply based on counting visits to each observation, while for continuous environments, we use a grid to collect visitation statistics. The relative novelty weight $w_N$ for selecting a cell is computed by
\begin{equation}
    w_N(\text{cell}) = \frac{1}{\left(p_v(\text{cell}) + \epsilon\right)^n}
\end{equation}
where $p_v(\text{cell})$ is the number of steps in the cell in relation to total steps taken, $\epsilon$ is a small positive number ($0.01$), and $n$ is a decay factor guiding how heavily it should prioritize cells with low visitations counts. Grid methods scale poorly with observation size, but the environments in this study are simple enough.
Once a grid cell is selected, a goal point is selected by sampling uniformly within it.



\subsubsection*{Intermediate success rate selection}

With the intermediate success rate goal selection method, a grid of cells is mapped onto the observation space in the same way as for novelty selection. Instead of visitations, targeted and successful goals are tracked in each cell. The rationale here is that we should learn some goals decently well before moving on to others.
The relative weight $w_S$ for selecting a cell is defined similarly to novelty selection, with
\begin{equation}
    w_S(\text{cell}) = \frac{1}{\left(\left|p_s(\text{cell}) - p_{st}\right| + \epsilon\right)^n}
\end{equation}
where $p_s(\text{cell})$ is the success rate in each cell, $p_{st}$ is a target success rate hyperparameter, $\epsilon$ is a small positive number (0.01), and $n$ is a hyperparameter that governs the degree to which proximity to the target success rate should be emphasized.
This is complemented with randomly selecting goals among visited but not targeted cells, in proportion to how many cells have been visited but not targeted, as well as some uniform random selection over all observations.





\subsection*{Environments}


We have applied our method to three environments: Cliff Walking, Frozen Lake and Pathological Mountain Car, with trials on more environments being ongoing work. The former two are part of the OpenAI Gym framework \citep{brockman2016openaigym}, and the latter is an adaptation of Mountain Car from the same framework. All environments currently examined are fully observable with discrete actions. 

The Cliff Walker environment is the simplest, it is discrete and deterministic, but with a few unreachable states, providing a simple environment and testing how much of an issue unreachable states are. Frozen Lake has stochastic transitions, with different maximum achievable success rates for each state; this can be deceptive when success rates are used to guide goal selection. The Pathological Mountain Car has continuous states and observations, exposing questions on how precise goal evaluation should be, if there are drawbacks with goals constrained to points, and how well our method can handle environments where actions are note easily reversible.


\subsubsection*{Cliff Walking}


The Cliff Walking environment \citep{brockman2016openaigym} is small, with a discrete grid of observations (represented by a single number each), and most observations (falling of the cliff moves the agent to the starting state). Transitions are deterministic and can in most cases easily be reversed by performing the inverse action, with termination only on the environment goal. The external rewards are $-1$ for each step and $-100$ for falling off the cliff. We truncate episodes after $300$ steps, to allow for hindsight experience replay.

\subsubsection*{Frozen lake}
The Frozen Lake environment \citep{brockman2016openaigym} is similar to Cliff Walking, a discrete grid with locations to avoid. However, with default parameters, it has stochastic transitions. It has a $1/3$ chance of going in the intended direction and $1/3$ for each of the perpendicular ones, making its optimal policy less trivial. In this case pitfalls end the episode, instead of resetting to the start like in the Cliff Walker environment. It truncates after $100$ steps.

\subsubsection*{Pathological Mountain Car}
\begin{figure}[htbp]
    \centering
    \begin{subfigure}[b]{0.3\textwidth}
        \centering
        \includegraphics[width=\textwidth]{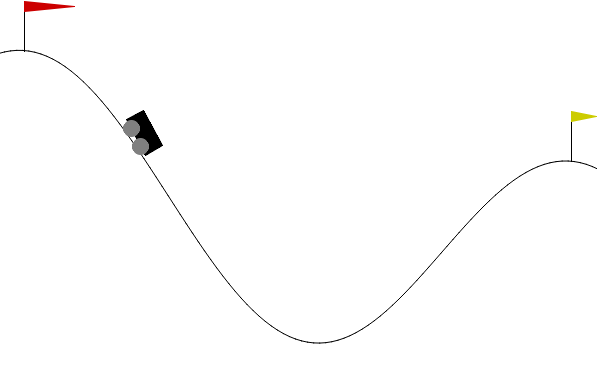}
        \caption{} 
        \label{fig:pmc_img}
    \end{subfigure}
    \hfill
    \begin{subfigure}[b]{0.5\textwidth}
        \centering
        \includegraphics[width=0.75\textwidth]{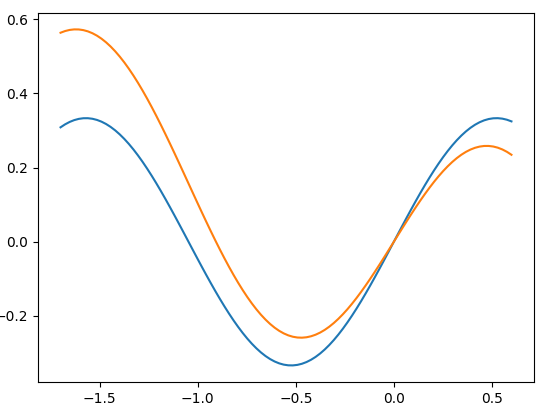}
        \caption{} 
        \label{fig:pmc_diff}
    \end{subfigure}
    \caption{The Pathological Mountain Car environment, visualized in \ref{fig:pmc_img}. An adaption of \cite{brockman2016openaigym}'s Mountain Car, with an additional goal, and a shift making one hill steeper (with higher external reward) and the other slightly flatter. \ref{fig:pmc_diff} shows the difference in inclination between our implementation (orange) and the original Mountain Car (blue).}
    \label{fig:pmc_env}
\end{figure}

We provide an adaption of \cite{brockman2016openaigym}'s Mountain Car, called Pathological Mountain Car, inspired by an environment of the same name introduced by \cite{chakrabortyDealingSparseRewards2023}. It differs from Open AI's Mountain Car by applying a small linear shift in height, making one hill harder to reach than the other, and placing a terminating state there, visualized in Figure \ref{fig:pmc_env}. The rewards are $500$ for reaching the summit of the tall hill, $10$ for the low hill. Unlike the regular Mountain Car environment, there is no penalty for each step. It truncates after $300$ steps. The Pathological Mountain car has the same continuous states and observations (horizontal position and velocity), and discrete actions, as the original. In order to reach the high-value goal, an agent would need to get quite near the low-value goal, which can make greedy agents miss the greater payout. In all mountain car environments, unlike the grid environments above, there is a lot of momentum in the system, so actions are not easily reversible.



\section*{Experiments and Results}



In the first section, our method is contrasted with a baseline RL algorithm using external rewards, and in the second section, we investigate the performance on different goals over time. Single standard deviations are included in all plots, and gaussian smoothing is applied for legibility. 

\subsection*{Comparison to external reward aware baseline} \label{sec:compare_to_baseline}

\begin{figure}[htbp]
    \centering
    \begin{subfigure}[b]{0.45\textwidth}
        \centering
        \includegraphics[width=\textwidth]{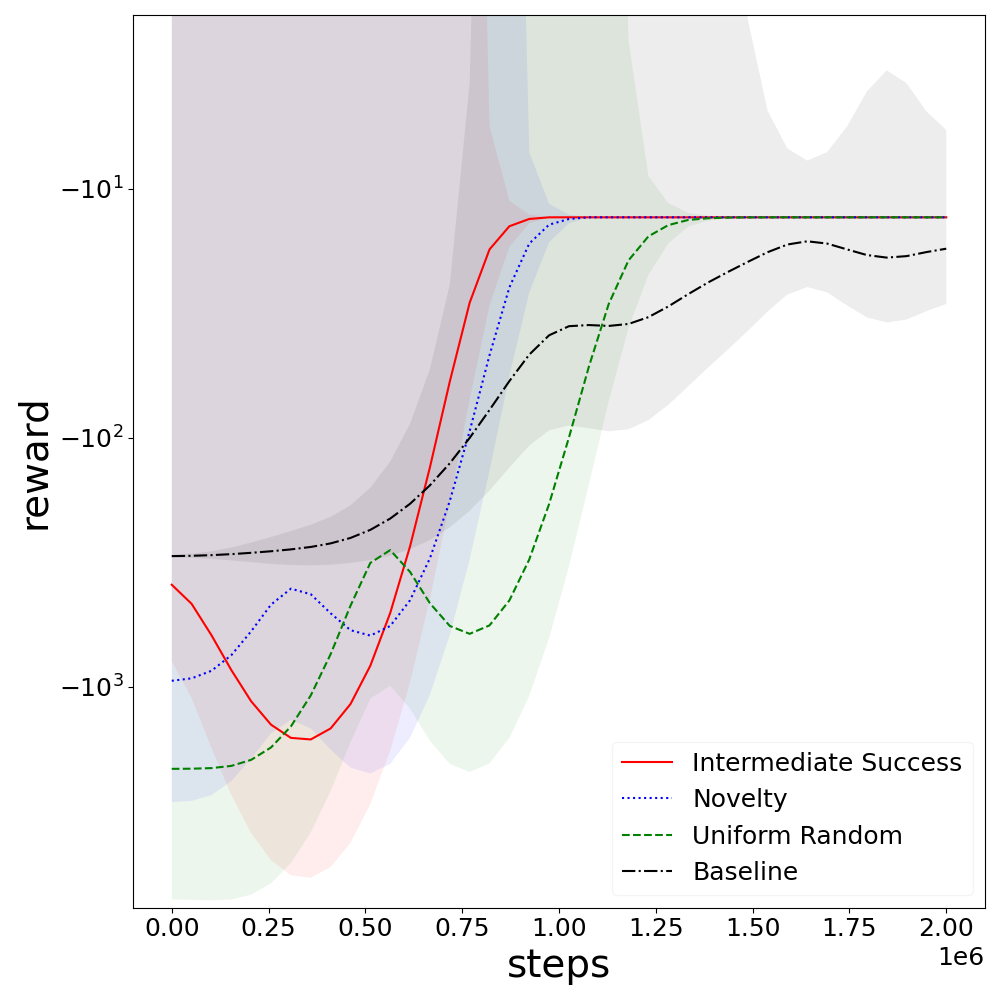}
        \caption{} 
        \label{fig:cliff_reward}
    \end{subfigure}
    \hfill
    \begin{subfigure}[b]{0.45\textwidth}
        \centering
        \includegraphics[width=\textwidth]{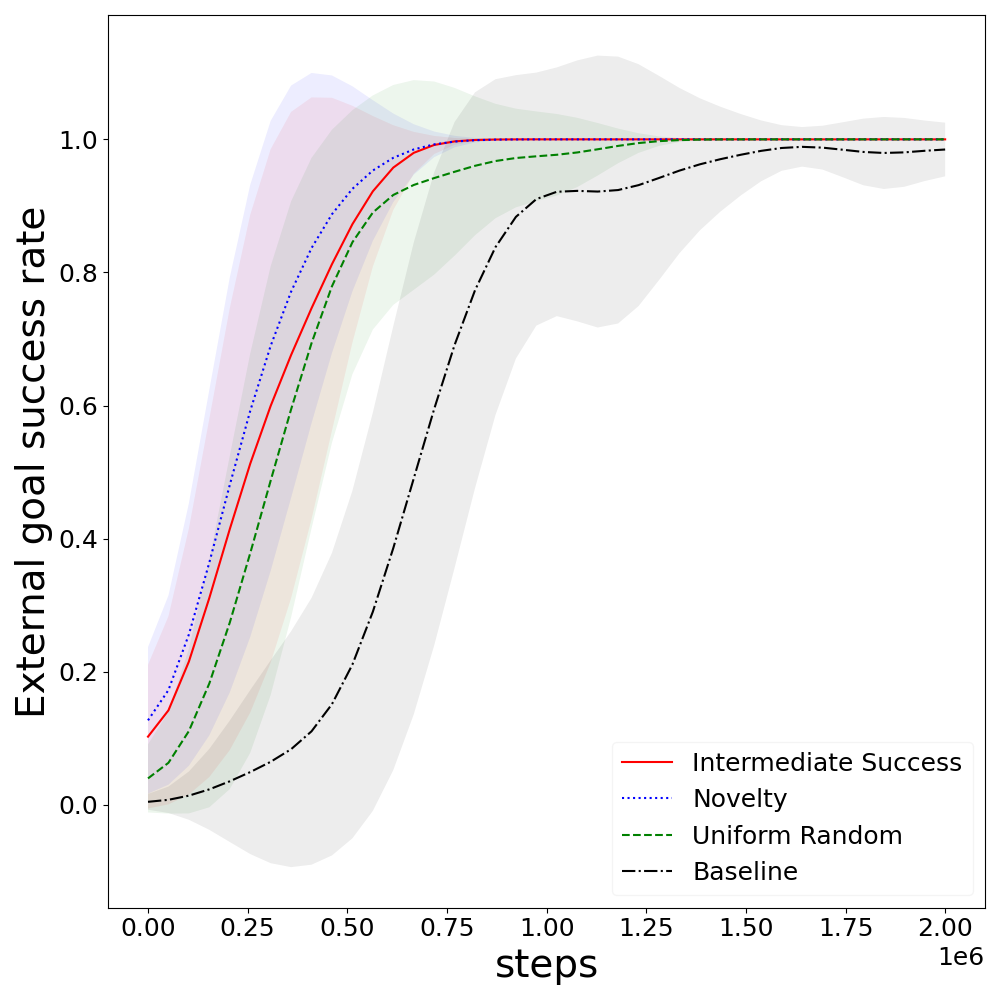}
        \caption{} 
        \label{fig:cliff_optimum}
    \end{subfigure}
    \caption{Cliff Walker environment, evaluation reward with symlog scale \ref{fig:cliff_reward}, as a function of training steps, and optimal behavior rate, \ref{fig:cliff_optimum}, with 8 experiments for each method. \textbf{Takeaway:} Our solution reaches the optimal policy quicker.
    }
    \label{fig:cliff_results}
\end{figure}

Compared to non-goal-conditioned reinforcement learning with DQN, our method reaches optimum faster on the Cliff Walker problem, as shown in Figure~\ref{fig:cliff_optimum}. Since our method does not see the external reward signal, it does not avoid cliffs as well in early training, seen in Figure~\ref{fig:cliff_reward}, but still stabilizes to the optimal solution slightly faster than the baseline with all goal selection methods. 

\begin{figure}[ht]
    \begin{center}
        \includegraphics[width=0.45\textwidth]{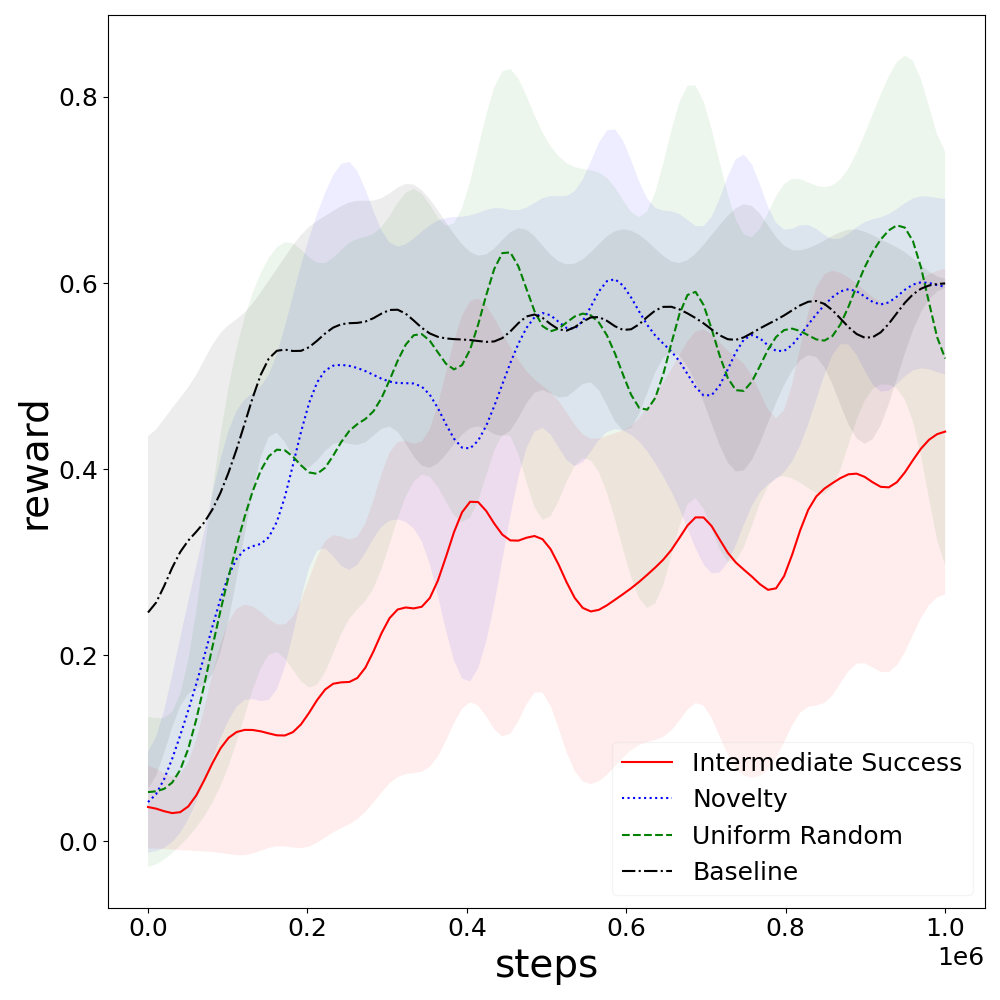}
    \end{center}

    \caption{Evaluation reward for Frozen Lake, as a function of training steps, with 4 experiments for each method. \textbf{Takeaway:} Our method gets comparable results, reward-free, with all but intermediate difficulty selection, but our methods and the baseline are worse than an oracle (with $\approx 0.7$ average reward).}
    \label{fig:frozen_reward}
\end{figure}

For the Frozen Lake environment, our method reaches a stable level of similar reward 2-3 times slower than the baseline, see Figure~\ref{fig:frozen_reward}, but neither reaches the optimal average reward (of about $0.7$) in the allotted training time. \
Intermediate difficulty selection performs worse on Frozen Lake. Since the environment is stochastic, the optimal success rate for several of the goals is less than $100\%$, and sometimes is similar to the target success rate. This leads to a goal selection bias, focusing training on such goals even when the agent already has the optimal policy for them.

\begin{figure}[htbp]
    \centering
    \begin{subfigure}[b]{0.45\textwidth}
        \centering
        \includegraphics[width=\textwidth]{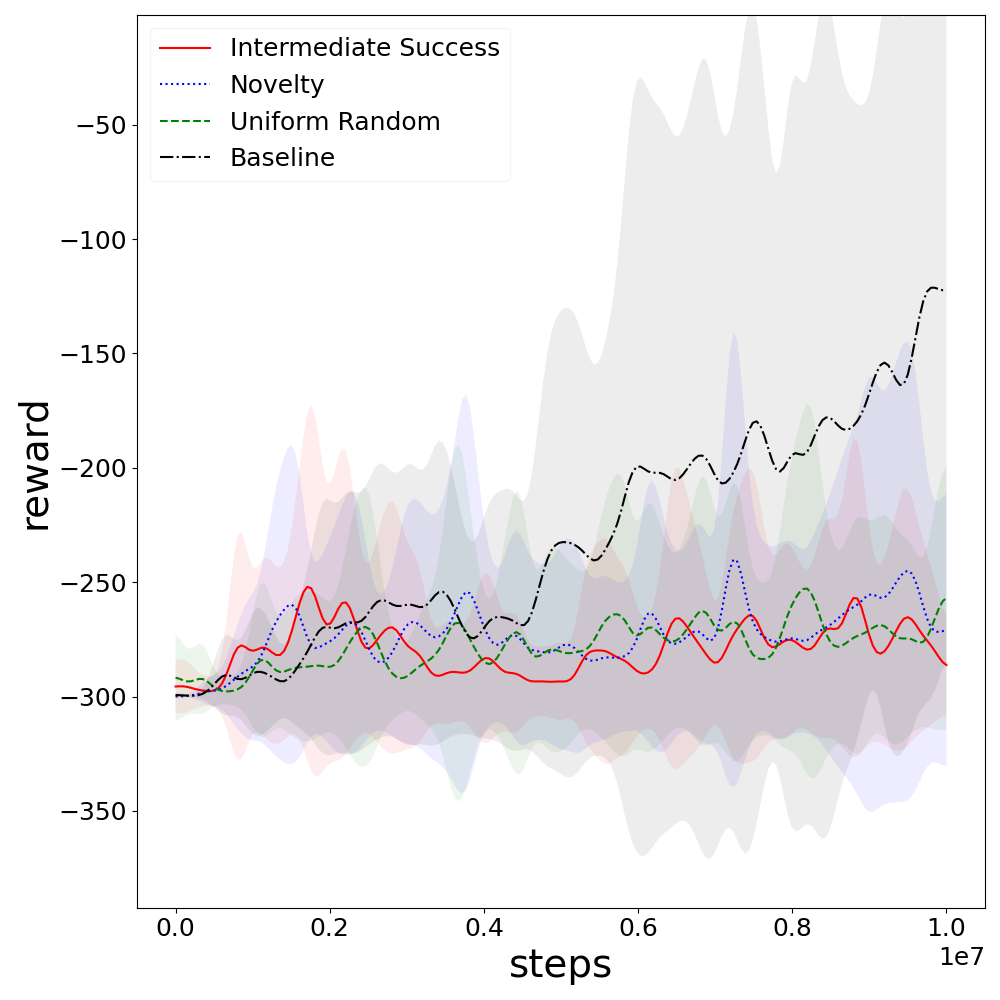}
        \caption{} 
        \label{fig:pmc_reward}
    \end{subfigure}
    \hfill
    \begin{subfigure}[b]{0.45\textwidth}
        \centering
        \includegraphics[width=\textwidth]{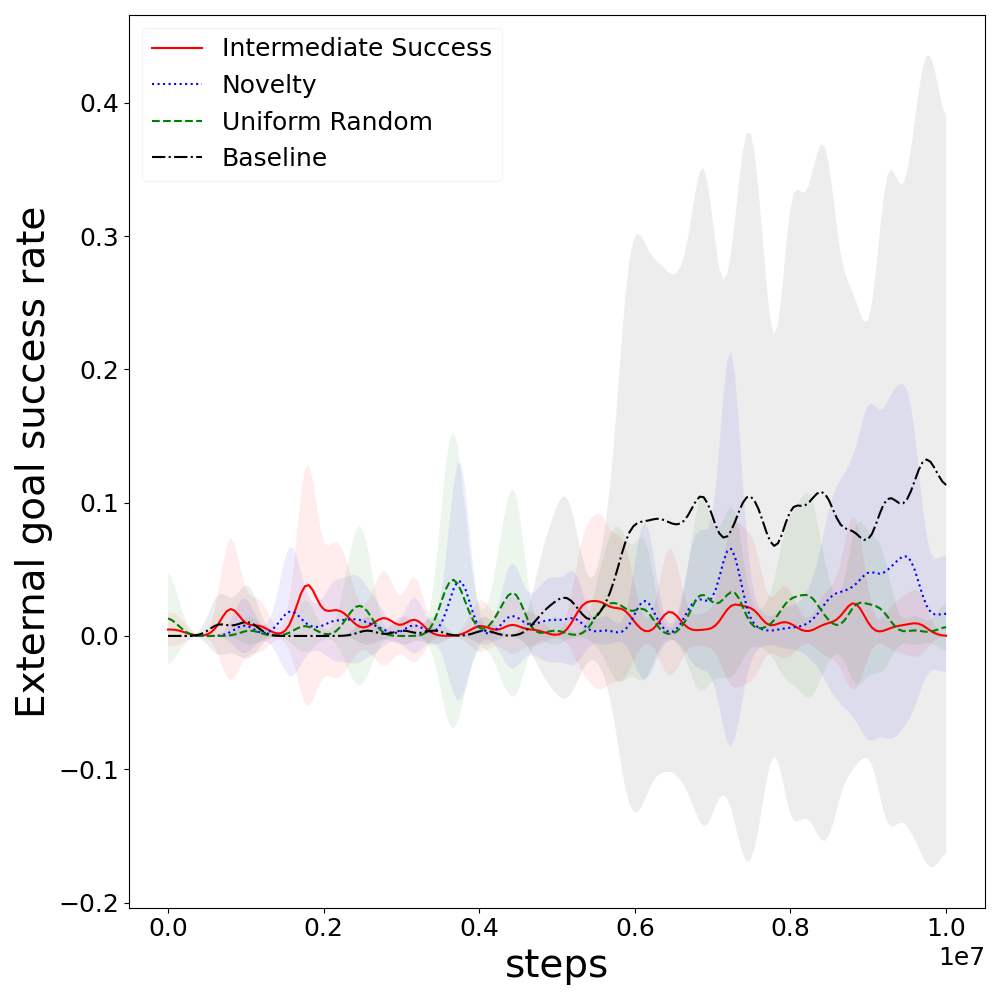}
        \caption{} 
        \label{fig:pmc_hardest}
    \end{subfigure}
    \caption{Evaluation reward, \ref{fig:pmc_reward}, and rate of success for reaching the hard goal, \ref{fig:pmc_hardest}, with 8 experiments for each method on Pathological Mountain Car, as a function of training steps. When evaluating goal methods, the hard hill is given as target goal. \textbf{Takeaway:} 
    Our method reaches the hard goal faster than the baseline, but does not retain the ability to reach it consistently.}
    \label{fig:pmc_results}
\end{figure}

The Pathological Mountain Car environment is more difficult to learn than the two earlier environments. As Figure~\ref{fig:pmc_results} shows, neither the baseline nor our method manage to reach the harder goal consistently. Our method reaches it intermittently, while the baseline reaches it in two out of eight experiments, and in one of those cases then learns to reach it more consistently. Experiments with a longer training time could determine which method is most consistent in learning to reach the hard goal.

This task exposes an issue with our method; our goals are observational points, specifying both a target position and a target velocity. The real goal here is to get the (horizontal) position to be less than $-1.6$, independent of velocity, and thus our method ignores many ways to complete the task and could even terminate prematurely if the validation goals are not set well. A dedicated solution could, of course, ignore or reformat the goal format to account for the particular environment, but we want to provide an environment-agnostic method. In ongoing work, we are expanding the goal formulation to take goal ranges. This would both allow evaluation with goals like the ones in this environment, and allow the goal selector to tune the goal difficulty by describing goals of varying scope, without leaving the environment-agnostic formulation.

\subsection*{Goal success rates}
\label{sec:goal_results}




Learning to reach any goal is a much broader task than external reward reinforcement. There is at least one optimal policy for each goal, instead of a global optimum for the entire environment. Agents trained with our method do not know about external rewards; instead, they are taught to reach all goals presented to them (as targeted goals, or hindsight goals during experience replay). To evaluate their performance, one should therefore track performance on a wide set of goals in the environment.

In the Cliff Walker environment, all goal selection methods converge quickly. The optimal policy is found for all goals at nearly the same time, and the average success rate for all goals reaches the expected $38/48 \approx 80\%$. It cannot reach $100\%$ since the ten cliff locations reset to the starting position and cannot be reached.

\begin{figure}[htbp]
    \centering
    \begin{subfigure}[b]{0.45\textwidth}
        \centering
        \includegraphics[width=\textwidth]{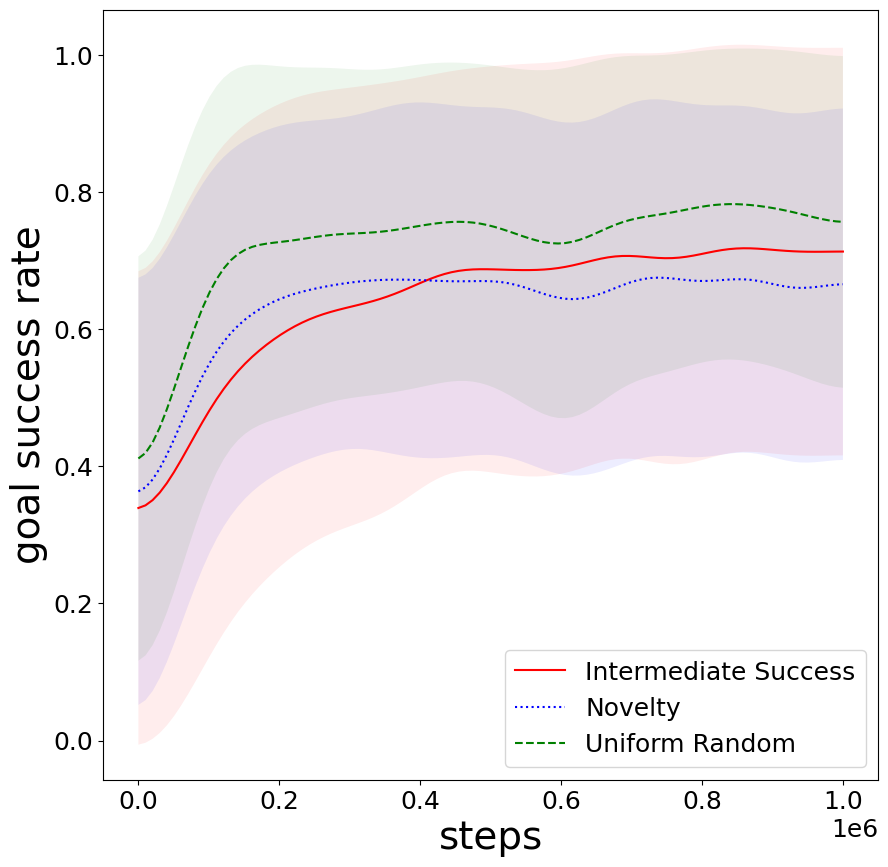}
        \caption{} 
        \label{fig:frozen_goal}
    \end{subfigure}
    \hfill
    \begin{subfigure}[b]{0.45\textwidth}
        \centering
        \includegraphics[width=\textwidth]{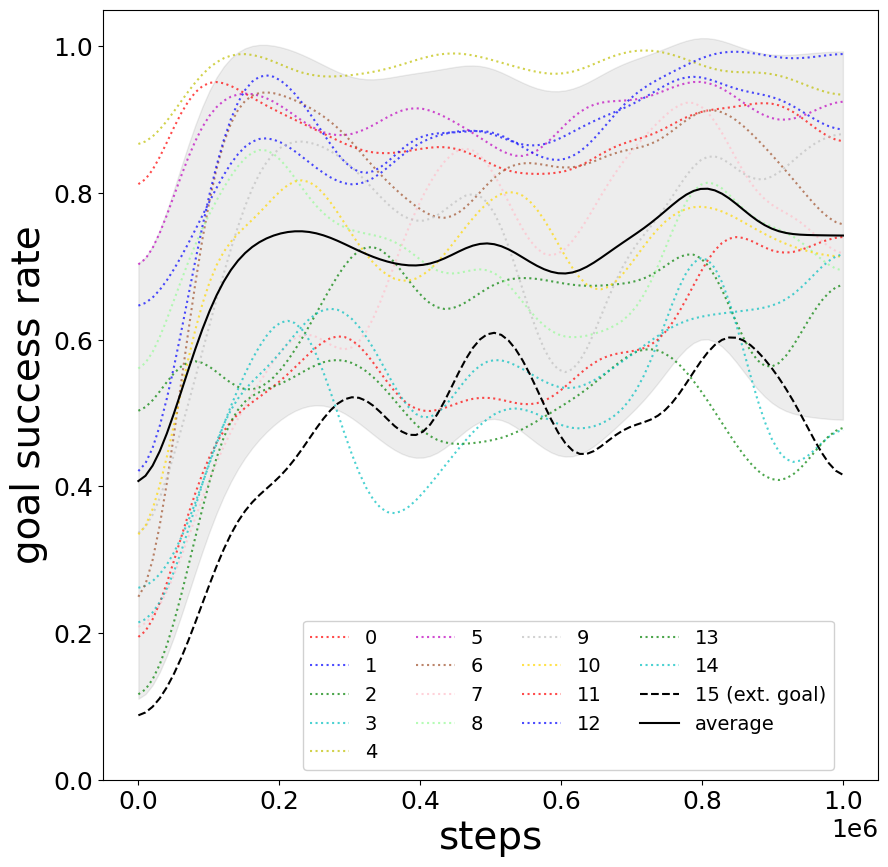}
        \caption{} 
        \label{fig:frozen_goals_in_one_run}
    \end{subfigure}
    \caption{The average goal success rate for our methods on Frozen Lake, \ref{fig:frozen_goal}, and an example of how success rates vary for goals when training an agent, \ref{fig:frozen_goals_in_one_run}. In \ref{fig:frozen_goals_in_one_run} goal 15 has the external reward (though our agent is unaware of that). \textbf{Takeaway:} The average stabilizes quickly, but individual goal performance fluctuates wildly, though this could in part be due to the stochasticity of the environment.}
    \label{fig:frozen_goal_results}
\end{figure}

The Frozen Lake environment is mainly characterized by its stochastic transitions, leading to extra uncertainty in both learning and evaluation. Despite this, the average goal performance is fairly stable, as shown in Figure~\ref{fig:frozen_goal}. When inspecting the training of a single agent, Figure~\ref{fig:frozen_goals_in_one_run}, one observes that performance on individual goals is unstable. If this is due to forgetting, the stochastic nature of the environment, or something else, is yet to be determined.

\begin{figure}[htbp]
    \centering
    \begin{subfigure}[b]{0.45\textwidth}
        \centering
        \includegraphics[width=\textwidth]{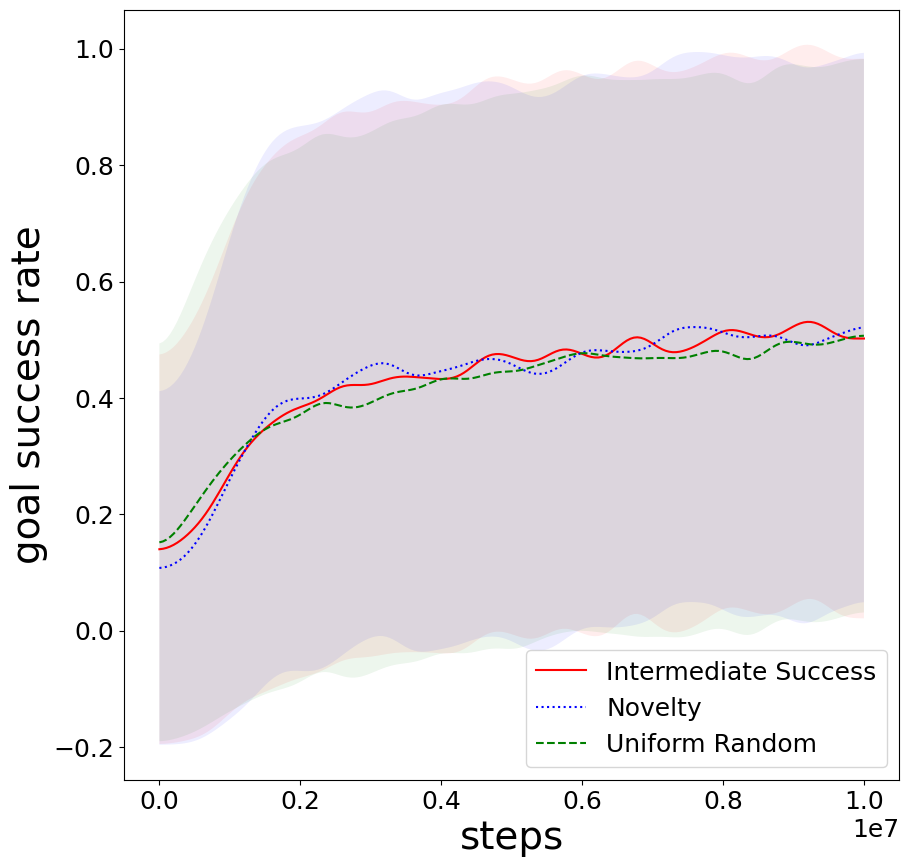}
        \caption{} 
        \label{fig:pmc_goal}
    \end{subfigure}
    \hfill
    \begin{subfigure}[b]{0.45\textwidth}
        \centering
        \includegraphics[width=\textwidth]{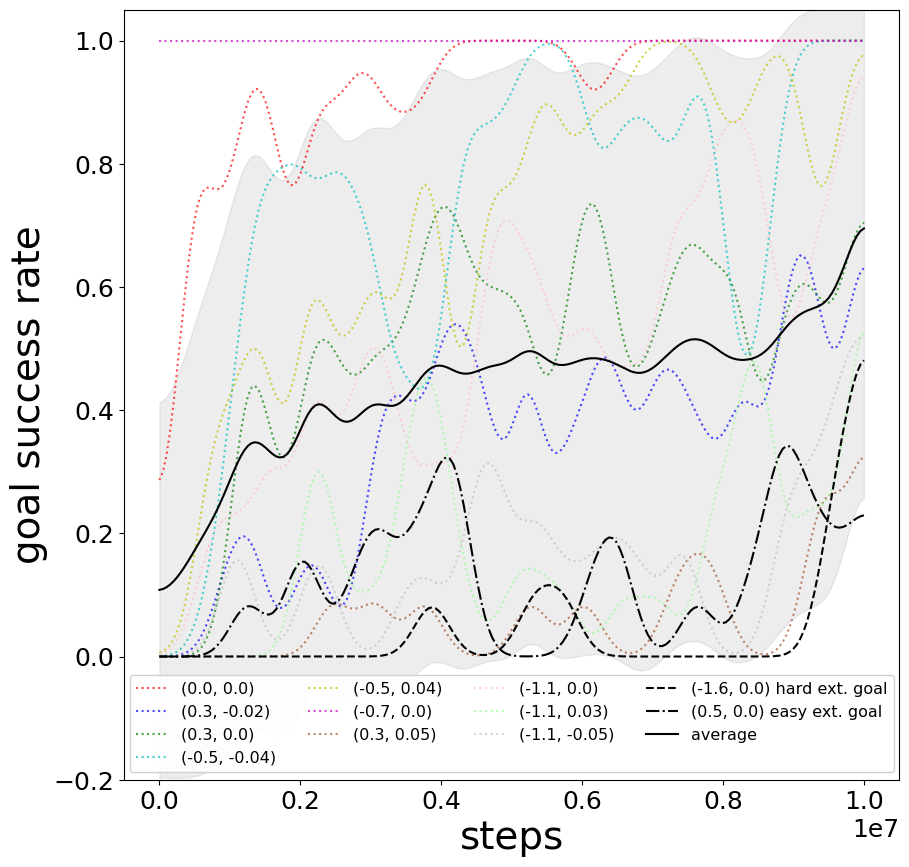}
        \caption{} 
        \label{fig:pmc_goals_in_one_run}
    \end{subfigure}
    \caption{The average goal success rate for our methods when trained on the Pathological Mountain Car environment, \ref{fig:pmc_goal}, and an example of how success rates vary for goals when training an agent, \ref{fig:pmc_goals_in_one_run}. In \ref{fig:pmc_goals_in_one_run} goal are described as $(x, \dot{x})$ where $x$ is horizontal position, and $\dot{x}$ is horizontal velocity, with $(-1.6, 0.0)$ and $(0.5, 0.0)$ corresponding to reaching the hard goal and easy goals respectively (while becoming stationary). \textbf{Takeaway:}
    The success rates for individual goals are unstable, even though the environment is deterministic, but the average success rate for all measured goals increases steadily.}
    \label{fig:pmc_goal_results}
\end{figure}

When studying goal performance in the Pathological Mountain Car environment, measurements cannot capture the goal performance on all goals, as it is an infinite set. A set of spread-out goals is selected as a sample of the overall goal success rate. Evaluations of this set during training, Figures~\ref{fig:pmc_goal}, \ref{fig:pmc_goals_in_one_run}, show that while the average goal success rate is similar for most goal selection methods, it is very unstable for individual goals. This is similar to the results for the Frozen Lake environment, but here the environment is deterministic, suggesting that forgetting or other training instability might be the cause. It should be noted that the average goal success rate is slowly growing, indicating that longer training time could be enough to achieve a consistent policy for all goals.

\section*{Conclusion}

In this paper, we have shown how goal conditioning can be used to learn reward-free in an environment-agnostic way, and we provide code to apply this method to non-goal environments. There are many open questions, but so far we can conclude that by autonomously selecting goals, our method can solve different tasks with accuracy and training times comparable to a non-goal conditioned externally reward guided baseline.


In ongoing and future work, we will explore how this method performs in more varied environments. 
We are interested in whether model-based methods can improve training efficiency by decoupling the learning of a transition model from the goal-based evaluation of states, observations, or actions. 
We also want to study the impact of more flexible goal formulation. 
A study on goal selection, its interaction with hindsight experience replay, and other ways to gather and use training data is ongoing. 
For complex and partially observed environments, other means of representing goals might be needed, and we are interested in exploring if goals can be selected within an embedding space, or if such changing representations are unsuitable.



\appendix




\subsubsection*{Acknowledgments}
\label{sec:ack}
Thank you Samuel Blad, for helping me find a bug that had evaded me for months.

This work was partially supported by the Wallenberg AI, Autonomous Systems and Software Program (WASP) funded by the Knut and Alice Wallenberg Foundation.



\bibliography{bibtex}

@article{andrychowiczHindsightExperienceReplay2018,
  title={Hindsight experience replay},
  author={Andrychowicz, Marcin and Wolski, Filip and Ray, Alex and Schneider, Jonas and Fong, Rachel and Welinder, Peter and McGrew, Bob and Tobin, Josh and Pieter Abbeel, OpenAI and Zaremba, Wojciech},
  journal={Advances in neural information processing systems},
  volume={30},
  year={2017}
}

@article{colasAutotelicAgentsIntrinsically2022,
  title = {Autotelic {{Agents}} with {{Intrinsically Motivated Goal-Conditioned Reinforcement Learning}}: {{A Short Survey}}},
  shorttitle = {Autotelic {{Agents}} with {{Intrinsically Motivated Goal-Conditioned Reinforcement Learning}}},
  author = {Colas, C{\'e}dric and Karch, Tristan and Sigaud, Olivier and Oudeyer, Pierre-Yves},
  year = {2022},
  month = jul,
  journal = {Journal of Artificial Intelligence Research},
  volume = {74},
  pages = {1159--1199},
  urldate = {2023-09-11},
  copyright = {Copyright (c)},
  langid = {english}
}

@inproceedings{colasCURIOUSIntrinsicallyMotivated2019,
  title={Curious: intrinsically motivated modular multi-goal reinforcement learning},
  author={Colas, C{\'e}dric and Fournier, Pierre and Chetouani, Mohamed and Sigaud, Olivier and Oudeyer, Pierre-Yves},
  booktitle={International conference on machine learning},
  pages={1331--1340},
  year={2019},
  organization={PMLR}
}

@article{ecoffetFirstReturnThen2021,
  title = {First Return, Then Explore},
  author = {Ecoffet, Adrien and Huizinga, Joost and Lehman, Joel and Stanley, Kenneth O. and Clune, Jeff},
  year = {2021},
  month = feb,
  journal = {Nature},
  volume = {590},
  number = {7847},
  pages = {580--586},
  publisher = {Nature Publishing Group},
  urldate = {2021-03-05},
  copyright = {2021 The Author(s), under exclusive licence to Springer Nature Limited},
  langid = {english}
}

@inproceedings{florensaAutomaticGoalGeneration2018,
  title={Automatic goal generation for reinforcement learning agents},
  author={Florensa, Carlos and Held, David and Geng, Xinyang and Abbeel, Pieter},
  booktitle={International conference on machine learning},
  pages={1515--1528},
  year={2018},
  organization={PMLR}
}

@inproceedings{liuGoalConditionedReinforcementLearning2022a,
  title = {Goal-{{Conditioned Reinforcement Learning}}: {{Problems}} and {{Solutions}}},
  shorttitle = {Goal-{{Conditioned Reinforcement Learning}}},
  booktitle = {Thirty-{{First International Joint Conference}} on {{Artificial Intelligence}}},
  author = {Liu, Minghuan and Zhu, Menghui and Zhang, Weinan},
  year = {2022},
  month = jul,
  volume = {6},
  pages = {5502--5511},
  urldate = {2024-05-29},
  langid = {english}
}

@inproceedings{pathakCuriosityDrivenExplorationSelfSupervised2017,
  title = {Curiosity-{{Driven Exploration}} by {{Self-Supervised Prediction}}},
  booktitle = {2017 {{IEEE Conference}} on {{Computer Vision}} and {{Pattern Recognition Workshops}} ({{CVPRW}})},
  author = {Pathak, Deepak and Agrawal, Pulkit and Efros, Alexei A. and Darrell, Trevor},
  year = {2017},
  month = jul,
  pages = {488--489},
  publisher = {IEEE},
  address = {Honolulu, HI, USA},
  urldate = {2019-04-22},
  langid = {english}
}

@article{stable-baselines3,
  author  = {Antonin Raffin and Ashley Hill and Adam Gleave and Anssi Kanervisto and Maximilian Ernestus and Noah Dormann},
  title   = {Stable-Baselines3: Reliable Reinforcement Learning Implementations},
  journal = {Journal of Machine Learning Research},
  year    = {2021},
  volume  = {22},
  number  = {268},
  pages   = {1-8},
  url     = {http://jmlr.org/papers/v22/20-1364.html}
}

@article{brockman2016openaigym,
  title={Openai gym},
  author={Brockman, Greg and Cheung, Vicki and Pettersson, Ludwig and Schneider, Jonas and Schulman, John and Tang, Jie and Zaremba, Wojciech},
  journal={arXiv preprint arXiv:1606.01540},
  year={2016}
}

@inproceedings{chakrabortyDealingSparseRewards2023,
  title = {Dealing with {{Sparse Rewards}} in {{Continuous Control Robotics}} via {{Heavy-Tailed Policy Optimization}}},
  booktitle = {2023 {{IEEE International Conference}} on {{Robotics}} and {{Automation}} ({{ICRA}})},
  author = {Chakraborty, Souradip and Bedi, Amrit Singh and Weerakoon, Kasun and Poddar, Prithvi and Koppel, Alec and Tokekar, Pratap and Manocha, Dinesh},
  year = {2023},
  month = may,
  pages = {989--995},
  urldate = {2023-12-12}
}

@article{mnih2013playingatarideepreinforcement,
  title={Playing atari with deep reinforcement learning},
  author={Mnih, Volodymyr and Kavukcuoglu, Koray and Silver, David and Graves, Alex and Antonoglou, Ioannis and Wierstra, Daan and Riedmiller, Martin},
  journal={arXiv preprint arXiv:1312.5602},
  year={2013}
}

@inproceedings{haarnoja2018softactorcriticoffpolicymaximum,
  title={Soft actor-critic: Off-policy maximum entropy deep reinforcement learning with a stochastic actor},
  author={Haarnoja, Tuomas and Zhou, Aurick and Abbeel, Pieter and Levine, Sergey},
  booktitle={International conference on machine learning},
  pages={1861--1870},
  year={2018},
  organization={PMLR}
}

@inproceedings{he2015deepresiduallearningimage,
  title={Deep residual learning for image recognition},
  author={He, Kaiming and Zhang, Xiangyu and Ren, Shaoqing and Sun, Jian},
  booktitle={2016 IEEE Conference on Computer Vision and Pattern Recognition (CVPR)},
  pages={770--778},
  year={2016}
}
\bibliographystyle{rlj}

\beginSupplementaryMaterials

\section*{Hyperparameters}

\HA{TODO add hyperparameter details}

For non-listed hyperparameters, Stable-Baselines3 defaults are used, but if we missed listing something, don't hesitate to reach out.

\subsection*{Goal evaluation}

In the presented evaluations, goals are considered successful if a normalized distance vector between the observation and the goal is less than $0.1$.

\subsection*{Goal selection}

For both novelty and intermediate success rate selection, $\epsilon = 0.01$. For data shown in this paper $n = 2$, but $n = 1$ produces similar results. The grid used to collect statistics for goal selection has $10\,000$ cells for Pathological Mountain Car, and matches obs space exactly for discrete environments.

With intermediate success rate selection, $0.9$ is the target success rate for CliffWalker, $0.75$ for FrozenLake and Pathological Mountain Car. 

In all these cases, $10\%$ of the time, a random goal is selected from the observation space.

\subsection*{Network}
Two Q-networks are used, one simple with 3 fully connected layers of 256 nodes each, and one resnet inspired network with 4 sequential residual nodes. Each residual node consists of 4 fully connected layers with 256 nodes, with the output of the final one being additive with the input of the residual node, similar to the original ResNet by \citet{he2015deepresiduallearningimage}. The simple network is used for the Frozen Lake environment and the ResNet inspired network for Cliff Walker and Pathological Mountain Car. We used Stable Baseline's implementation of Hindsight Experience Replay, adapted to handle termination when reaching goals.

Agents are trained with $0.001$ learning rate, gamma $0.95$, batch size $512$, and train frequency $512$. HER replay buffer uses "future" goal reselection, with $4$ hindsight goals for each use of unaltered experience. The replay buffer is $1\,000\,000$ steps long.

Cliffwalker was trained for $2\,000\,000$ steps, FrozenLake $1\,000\,000$ steps and Pathological mountain car $10\,000\,000$ steps.

\end{document}